\documentclass[journal]{IEEEtran}
\usepackage{blindtext}
\usepackage{graphicx}

\usepackage{cite}

\usepackage[cmex10]{amsmath}

\usepackage{amsfonts}

\usepackage{algorithm}
\usepackage{algorithmic}

\usepackage{subcaption}

\usepackage{float}
\floatstyle{plaintop}
\restylefloat{table}

\usepackage{url}
\usepackage{hyperref}

\begin{document}

\title{Learning to Play General Video-Games via an Object Embedding Network}

\author{\textbf{William Woof} \& \textbf{Ke Chen}

School of Computer Science, The University of Manchester

%Oxford Road, Manchester, UK, M13 9PL 

Email: \{william.woof, ke.chen\}\url{@manchester.ac.uk}} %TODO: Add authors and

%\markboth{Journal of \LaTeX\ Class Files,~Vol.~6, No.~1, January~2007}%
%{Shell \MakeLowercase{\textit{et al.}}: Bare Demo of IEEEtran.cls for Journals}



\maketitle

\begin{abstract}
%\boldmath

Deep reinforcement learning (DRL) has proven to be an effective tool for creating general video-game AI. However most current DRL video-game agents learn end-to-end from the video-output of the game, which is superfluous for many applications and creates a number of additional problems. More importantly,
directly working on pixel-based raw video data is substantially distinct
from what a human player does.
In this paper, we present a novel method which enables DRL agents to learn directly from object information.
This is obtained via use of an object embedding network (OEN) that compresses a set of object feature vectors of different lengths into a single fixed-length unified feature vector representing the current game-state and fulfills the DRL simultaneously.
We evaluate our OEN-based DRL agent by comparing to several state-of-the-art approaches on a selection of games from the GVG-AI Competition. Experimental results suggest that our object-based DRL agent yields performance comparable to that of those approaches used in our comparative study.

\end{abstract}

%\begin{IEEEkeywords}
%\end{IEEEkeywords}

\IEEEpeerreviewmaketitle

\section{Introduction}

%TODO: Could probably use some diagram on the first or second page to keep readers attention.

\emph{General video-game AI} (GVG-AI) is an area regarding the development of general algorithms that enable AI agents to play a wide range of different video-games with minimal tailoring to specific games.
While developing techniques for General AI is a key focus of research into GVG-AI, general video-game playing agents also have a number of applications within the games industry.
Asides from the obvious applications, such as a replacement to hand-coded in-game AI, GVG-AI can also either be used as a development tool or as a proxy for human play-testers. Such agents could be employed effectively in a wide array of applications from testing game balance \cite{cityconquest} to evaluating procedurally generated content \cite{nielsen2015general}.
As well as applications within games and games design, GVG-AI also has wider implications for the field of AI, as techniques which work well on video-games can often also be applied to real-world problems.

One promising area in the search for general video-game players is Deep Reinforcement Learning (DRL). DRL agents have been successfully applied to a wide range of video-games ranging from 2D arcade games \cite{mnih2015human} to challenging 3D shooters \cite{lample2017playing}.
These agents learn through interacting with the game autonomously, using a deep neural network to select actions based on the current state of the game. During this process the agent receives rewards (usually dictated by the in-game score) which indicate how well it is performing. By using an appropriate reinforcement learning algorithm the agent is able to modify its neural network in order to maximise this reward signal.
% Want to automate the agent as much as possible so methods that work generally are important
% The end-user is unlikely to be a ML expert
% Interpretability is another big issue
However, such agents are far from perfect, and can sometimes be difficult to apply in practice -- a problem compounded by the fact that they typically take a long time to train. A significant consideration into the design of these agents is how information from the game is presented to their neural networks (\emph{i.e.} the representation given to the agent), as well as the design of the networks themselves.

Current state-of-the-art DRL approaches to video-games learn directly from raw video data, using deep \emph{convolutional neural networks} (CNNs) \cite{mnih2015human}. While widely applicable, this approach is subject to limitations for certain applications.
%Learning from video data has a number of practical issues, especially when considering games-development applications.
For example, many games (e.g.,  \emph{Starcraft}) feature controllable cameras, meaning much of the game-state is obscured from the agent at any given point. Working around this, e.g.,  by putting the camera under the agent's control, adds additional complexity to the agent.
Additionally, in many cases it may be undesirable, or even impossible to produce a video-output for the agent to consume. Rendering videos for multiple agents may be prohibitively expensive, and in some cases there may be no obvious way to produce a good visual representation for NPCs (non-player characters).
More importantly, interpreting raw video data at a pixel level is substantially different from how human players appear to play, as studied in \cite{dubey2018investigating}. % And DQN uses objects anyway
% \cite{rosenfeld2018bridging}

Unlike many other reinforcement learning tasks, the ground-truth information about the current state of the environment is often available in video-games, although
such information needs to be organised and presented to an agent in some way. Hence, the use of this direct information about the current game-state could be alternative to working directly on raw video data. For instance, Samothrakis et al \cite{samothrakis2015neuroevolution} employ a fixed set of general features, i.e., distance to the nearest enemy, number of tokens collected and so on, to encapsulate the current game-state. While this approach often works well, those general features have to be handcrafted, which is laborious and requires human expertise. Moreover, this approach is relatively game-specific and hence generally inappropriate to GVG-AI.
To overcome this limitation, various game-independent object representations, have been employed  \cite{kunanusont2017general,narasimhan2017deep}.
In an object representation, each game-state observation is given as a list of objects, and their classes and attributes.
For example, the state of one round of the game \emph{Pong} might be represented by two objects of the class \texttt{bat}, with attributes of \texttt{x-coord}, \texttt{y-coord}, and \texttt{player}, and an object of the class \texttt{ball} with attributes \texttt{x-coord}, \texttt{y-coord}, \texttt{x-velocity}, and \texttt{y-velocity}.
Many video-games rely on objects for their internal representation of the game-state. For instance, the popular \emph{Unity 3D} game engine relies heavily on game objects, and even the early \emph{Atari} console used a primitive sprite-based system. In general, the use of object representations not only leads to an effective approach to representing game-states across a wide variety of video games but also has a number of practical benefits. For example, it allows the use of different subsets of objects for different agents.
Also, objects provide useful anchor points for applying various advanced reinforcement learning techniques such as hierarchical reinforcement learning \cite{topin2015portable}, intrinsic motivation \cite{kulkarni2016hierarchical}, and planning \cite{garnelo2016towards}. %van2017hybrid
Given there are a different number of objects in different states of game, however, how to structure this information in a way that can be input into a conventional deep neural network is a key issue of using object representations with DRL.
Previous solutions to this problem mimic an image representation by using an ``object perception grid", where objects are overlayed onto a grid and mapped to the nearest cell determined by their $\texttt{x}$ and $\texttt{y}$ co-ordinates within the game. The number of objects mapped to each cell is then used as an input for a conventional neural network (either fully-connected or convolutional). Unfortunately, such a solution requires selecting an appropriate grid size manually, and entails a large input space, increasing the required neural network complexity. In general, it also still suffers from many of the same problems as encountered by using raw image representations, such as a restricted field of view.

In this paper, we present a novel approach to address the object representation issues in GVG-AI. To overcome all the aforementioned limitations, we adopt a specific type of neural network architecture, set networks \cite{zaheer2017deep,santoro2017simple,watters2017visual}, to develop our \emph{object embedding network} (OEN). This network can not only take a list of object-feature vectors of arbitrary lengths as input to produce just a single, yet unified, fixed-length representation of all the objects within the current game-state, but also be trained on a given task simultaneously. Hence, our OEN-based approach provides an alternative way to apply DRL algorithms within video-games, based on object information.
Our approach is generally motivated by recent advances within approaches to relational reasoning \cite{santoro2017simple} and dynamics prediction \cite{watters2017visual}, which suggest that working with objects, rather than raw data, can help scale up deep learning to more complicated tasks in a similar fashion to human information processing.

Our main contributions in this paper are summarised as follows:
\begin{enumerate}
    \item We propose an OEN model, based on set networks, for learning directly from sets of object feature vectors.
    \item We develop an OEN-based GVG-AI agent for playing general video games.
    \item We evaluate our approach on selected games from the GVG-AI competition and demonstrate that it performs comparably to a variety of other popular approaches for representing game states.
\end{enumerate}

The rest of this paper is organised as follows. Section \ref{sec:background} reviews related work. Section \ref{sec:OEN} presents our OEN model. Section \ref{sec:method} describes our object-based approach to GVG-AI. Section \ref{sec:experiments} describes our experimental settings and reports experimental results. Finally, Section \ref{sec:discussion} discusses issues and implications arising from this study.

%==================================================

\section{Related work}\label{sec:background}

\subsection{Deep Reinforcement Learning for video-games}

Reinforcement learning is a sub-field of machine learning where agents autonomously interact with an environment and seek to maximise some reward signal.
\emph{Deep reinforcement learning} (DRL) is an extension of classical `tabular' approaches to reinforcement learning which enable these techniques to be applied to more complicated problems.
By using the in-game score counter as a reward signal, DRL can be applied to develop agents for playing video-games \cite{mnih2015human}, making video-games a popular test-bed for new DRL algorithms.
Those algorithms work by using a deep neural network to `score' possible actions given a particular game-state, which is then trained according to a loss function. Such a loss function for DRL may be formulated based on the agent's past experience as well as the reward obtained.
%Rather than storing the agent's policy (or value function) directly in a look-up table, DRL algorithms rely on some deep neural network to approximate this function over the input state-space. This network can then be trained according to some loss function defined by the particular DRL algorithm.

The \emph{deep-Q network} (DQN) algorithm \cite{mnih2015human} is a pioneering work in applying DRL to general video games playing, where the DQN was trained to play a variety of games for the \emph{Atari 2600} games console. In the original DQN algorithm, the game-state is presented to the agent as a series of four images from four consecutive frames of video output, which is interpreted by a deep \emph{convolutional neural network} (CNN) with four input channels.
The success of this DQN algorithm has led to a number of alternative DRL algorithms for video-game playing  \cite{mnih2016asynchronous,pritzel2017neural} for the \emph{Atari} system, and these algorithms have also been applied to a variety of other video games \cite{lample2017playing,tessler2017deep}. While the DQN algorithm and its variants can be easily adapted to a variety of input types and network architectures, these mainly use the same input format; i.e. a sequence of frames from a game video stream. A notable exception is the use of object perception grids, which is described below in Section~\ref{sec:OORL}.

\subsection{Object-oriented reinforcement learning}\label{sec:OORL}

Reinforcement learning from objects has previously been studied within object-oriented reinforcement learning \cite{diuk2008object}, which allows for exploiting structural information at an object level. In object-oriented reinforcement learning, the state-space is expressed in terms of a set of objects. These objects all belong to some class from a fixed set of classes. Each object is an ordered tuple of object attributes, where the domain of these attributes is determined by the object class. For example, for a simple empty $5 \times 5$ grid-world, we might formulate the state-space with a single class, $\mathtt{Agent}$, of attributes $\mathtt{x}$ and $\mathtt{y}$, $\emph{Dom} (\mathtt{x}) = \emph{Dom} (\mathtt{y}) = \{ 1, 2, \dots , 5\}$. Then the initial state $s_0$ would be a single object $\mathtt{agent}$ from the $\mathtt{Agent}$ class with attribute values $\mathtt{x} = \mathtt{y} = 0$.

Conventional approaches to solving these problems usually involve planning algorithms, and first order logic \cite{hershkowitz2015learning}, or otherwise rely on the discrete nature of the environment, which is generally incompatible with DRL approaches. In particular, structuring this information in a way that can be used by a neural network is a challenging problem.
Very recently, this problem has been addressed via an object perception grid representation, e.g., \cite{kunanusont2017general,narasimhan2017deep}.
In this representation, objects are mapped onto grid squares, based on their \texttt{x} and \texttt{y} attributes. Each of these grid squares is then treated as an input neuron, which is set to $1$ if an object of a given class is present at that cell, and $0$ otherwise.
The full representation is then given by multiple input grids, one for each possible object class.
This effectively produces an image-like representation, which can be fed into a CNN. Apart from removing some of the complexity of the input, this approach suffers from most of the same problems as a visual representation. Moreover, this representation also requires the designer to select a grid size and coarsity. While many games may feature a natural choice for these, for many games it may require careful selection and additional tuning to select these parameters appropriately.
Additionally this process can only be applied where objects are related by a clear 2D structure.

\section{Object Embedding Network}\label{sec:OEN}

In this section, we present an \emph{object embedding network} (OEN) to learn a unified object representation from arbitrary sets of objects characterised by a variety of object features without being limited to 2D spatial structures. Our OEN model is based on an emerging class of deep neural-network architecture, set networks \cite{zaheer2017deep}, which were recently developed to tackle the input data in a set form.
In general, objects in a game-state naturally stand in a set form. By using the same principles behind set networks, our OEN transforms an arbitrary number of object feature vectors corresponding to a game state into a single fixed-length ``\emph{unified}" object representation.
This unified representation can be used for a variety of different purposes. Our OEN can be trained to learn a unified representation and fulfil a specific learning objective simultaneously.

At game-state $s_t$, assume that there is a set of objects, $O^{(t)}=\{ o_k^{(t)} \}_{k \in 1,\cdots, K_t}$, where each object, $o_k^{(t)}$, can be characterised by a feature vector (or a number of attributes),  $\pmb{x}_k^{(t)}$. Hence, the feature vectors of all $K_t$ objects collectively form a set,  $X^{(t)} =\{\pmb{x}_k^{(t)}\}_{k \in 1,\cdots, K_t}$, for game-state $s_t$. Our problem is how to learn a fixed-length unified feature vector that retain as much representative information conveyed by $K_t$ objects as possible for arbitrary $K_t$.

A common way to get representative information of a set of vectors is to compute some statistic about the set. In practice, this can be achieved using simple arithmetic pooling functions, e.g., max or sum pooling, applied element-wise, which condense an input set of vectors into a single fixed-length vector of the same dimension.
However, simply applying simple pooling functions over a set of object feature vectors is likely to incur a loss of important information. For example, if the object features consist of \texttt{x} and \texttt{y} co-ordinates then taking the mean of all object feature vectors simply ends up with the average position of all objects. While this is useful information, it does not convey the important information, e.g., ``is object $o_i^{(t)}$ next to object $o_j^{(t)}$?". Hereinafter, we drop out the explicit game-state index, $t$, to facilitate our presentation.

Motivated by set networks \cite{zaheer2017deep}, we deal with this problem by embedding raw feature vectors of objects into a higher dimensional space, which allows for retaining non-trivial information after pooling.
This can be achieved by applying a proper ``embedding" function $E$ to the feature vector of each object:
$$ E(X) := \{ E(\pmb x_1), ..., E(\pmb x_K) \}.$$
Let $\Pi$ to denote a pooling function, a unified representation of the object feature set, $X$, is then achieved by
$$\pmb r(X) = \Pi_{k \in 1,\cdots,K} E(\pmb x_k). $$

However, finding a proper embedding function explicitly is extremely difficult in general. Also, the optimal choice of such an embedding function is task-dependent. Instead of using an explicit embedding function, we can employ a neural network to learn an optimal embedding function. Furthermore, for a specific task based on the unified representation, we can incorporate another neural network for fulfilling the given learning objective and learning the optimal embedding function simultaneously.

\begin{figure}[th]
    \centering
    \includegraphics[width=4cm]{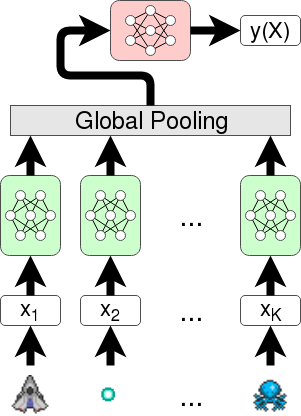}
    \caption{Object embedding network architecture.}
    \label{fig:objectembedding}
%\vspace*{-5mm}
\end{figure}

As illustrated in Fig. 1, a generic object embedding network (OEN) consists of \emph{embedding network} (shown in green), \emph{global pooling function} and \emph{task network} (shown in pink). For a set of $K$ objects in a game-state, $K$ identical embedding networks $E(X; \theta_E)$ are employed for embedding different objects, respectively, where $\theta_E$ is a collective notation of parameters shared by all $K$ embedding networks.
The global pooling function condenses the embedding object representations produced by $K$ embedding networks to yield a fixed-length unified object representation: $\pmb r(X) = \Pi_{k \in 1,\cdots,K} E(\pmb x_k, \theta_E)$.
Then, the unified representation is fed to the task network denoted by
$P(\pmb r(X) ; \theta_P)$ where $\theta_P$ is a collective notation of parameters in the task network.

Parameter estimation in the OEN is done by optimising a loss function defined on training data, $\mathcal{D}$, given for a specific task, $L(\mathcal{D}; \theta_E, \theta_P)$:
 $$
 \{ \theta_E^*, \theta_P^* \}= {\rm argmin}_{\theta_E, \theta_P}  L(\mathcal{D}; \theta_E, \theta_P).
 $$
For some loss function $L$, e.g., in a supervised learning task, a prediction-error based loss function can be used. In Section~\ref{sec:method}, we detail a loss function defined on transitions $(s_t, a_t, r_t, s_{t+1})$, drawn from experience-replay, for our reinforcement learning tasks.

Contextual information regarding the relationship between an object and other co-occurring ones at the same game-state can  play an important role. We can exploit this by replacing our embedding function $E(\pmb x)$ with a ``contextual" embedding function $E(\pmb x, X)$ which takes into account information from the wider set when embedding each object.
To this end, different techniques have been proposed in set networks, e.g., \cite{qi2016pointnet,zaheer2017deep,watters2017visual}, to explore this contextual information. 
Motivated by the work of \cite{qi2016pointnet,zaheer2017deep}, we adopt a simple global-context based method to explore the contextual information in our work. In this method, some statistic $\Pi$ of feature vectors of all the objects in the set $X$ is first estimated by $\hat{\pmb x}= \Pi_{\pmb x \in X} {\pmb x}$. Then, the feature vector of each object, $\pmb x$, is concatenated with this statistic vector, $\bar{\pmb x}$, to form a ``contextualised" feature vector of the object: $({\pmb x}, \hat{\pmb x})$. Instead of the feature vector of each object, $\pmb x$, its contextualised feature vector, $({\pmb x}, \hat{\pmb x})$, is fed to the embedding network in the OEN. In particular we adopt the same ``equivariant" transformation proposed in \cite{zaheer2017deep} which is given by:
 $$
 f_{equiv}(\pmb x, X) = \pmb x - maxpool(X).
 $$
By using this method, our OEN model can be extended to explore contextual information without altering its general architecture and learning algorithms.

%==================================================

\section{Model Description}
\label{sec:method}

In this section, we present our method for establishing an OEN-based DRL agent for playing general video games. We first describe object and feature extraction required by the OEN and then propose our OEN implementation and its deep Q-learning algorithm.

\subsection{Object and feature extraction}

For our agent, the process of identification and extraction of objects is  handled by the environment. That is, we assume that object extraction can be done directly via access to the ground truth of the environment. Hence, the wide-spread use of object-oriented programming languages should help with this process, as many objects are likely to be treated as such in code. Thus, the description of a game state is given in an object-oriented format; i.e., observation $O$ from a game-state is given as a list of objects, $O = \{ o_1, \cdots, o_K \}$.

Given a list of objects in this format, we still need to characterise those objects via a number of attributes to meet our requirement of our OEN-based DRL. In essence, this is a feature extraction process to obtain object feature vectors from the raw objects basted on their attributes, which leads to a pre-processing function \texttt{process\_observation} required by our OEN.
%Can the following  be curtailed?
A natural solution to feature extraction is concatenating all real-valued attributes of an object into a single feature vector. However, this solution results in a problem; while our OEN can handle only fixed-length object feature vectors, the number of attributes used to characterise an object is not fixed and different objects could have a different number and types of attributes. Hence, the user must select a set of attributes applicable to all the objects for a fixed-length feature vector. In this manner, a set of fixed-length feature vectors, $X=\{ \pmb x_1, \cdots, \pmb x_K \}$, can be extracted for a list of objects, $O$, in a game-state. This set of feature vectors are fed into our OEN, which acts as a value-network for the agent.

\begin{figure}[t]
    \centering
    \includegraphics[width=8cm]{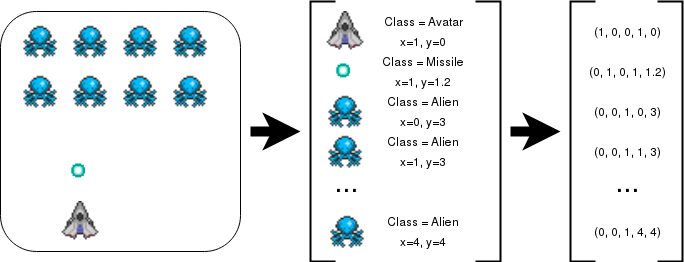}
    \caption{Exemplar object feature extraction process.}
    \label{fig:objrep}
    \vspace*{-7mm}
\end{figure}

Fig. \ref{fig:objrep} depicts an exemplar object feature extraction process. As seen in Fig. \ref{fig:objrep}, a game-state is broken down into a (finite) set of objects, along with a number of attributes for each object, as chosen by the user, e.g., position, class of object, and so on. This list of objects is then converted into a list of feature vectors by mapping each object to a fixed-length vector based on its attributes. In this example, a one-hot vector of the object class concatenated with the object's co-ordinates yields a 5-dimensional object feature vector.

\begin{figure*}[th]
    \centering
    \includegraphics[width=10cm]{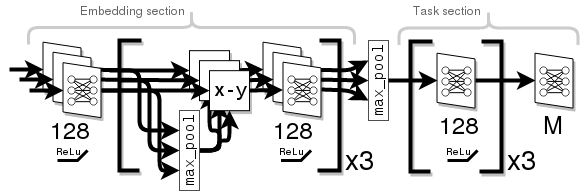}
    \caption{The object embedding network used to implement our DRL agent. }
    \label{fig:OEN}
    \vspace*{-7mm}
\end{figure*}

\begin{algorithm}
\caption{Q-learning algorithm for OEN}
\label{alg:objdqn}
\begin{algorithmic}
\STATE // \emph{Initialise agent}
\STATE initialise empty replay memory $M$
\STATE initialise $\theta$ for OEN Q-function $Q(s,a; \theta)$
\STATE $\theta_{target} \gets \theta$
\STATE // \emph{Initialise environment}
\STATE \texttt{env.reset()}
\STATE \texttt{step} $\gets 0$
\WHILE{\texttt{step} $\leq$ \texttt{max\_step}}
    \STATE \texttt{step} $\gets$ \texttt{step} $+1$
    \STATE // \emph{Get action from Q-function}
    \STATE $s_t \gets$ \texttt{env.get\_state()}
    \STATE $O_t \gets$ \texttt{process\_observation($s_t$)}
    %\STATE $a_t \gets argmax_a (Q(O^t,a; \theta))$
    \STATE $a_t \gets \texttt{epsilon\_greedy(}Q(O^t,a; \theta)\texttt{)}$
    %\STATE // \emph{Select random action with probability epsilon}
    %\IF {$rand() < \epsilon$}
    %    \STATE $a_t \gets$ \texttt{env.get\_random\_action()}
    %\ENDIF
    \STATE // \emph{Step environment and observe result}
    \STATE $r_t, s_{t'} \gets$ \texttt{env.apply\_action($a_t$)}
    %\STATE $s_{t'} \gets$ \texttt{env.get\_state()}
    \STATE $O_{t'} \gets$ \texttt{process\_observation($s_{t'}$)}
    \STATE $T_{t'} \gets$ \texttt{env.has\_ended()}
    \STATE add tuple $(O_t, a_t, r_t, O_{t'}, T_{t'})$ to $M$
    \STATE // \emph{Train network}
    \STATE sample tuple $({O'}_t, {a'}_t, {r'}_t, {O'}_{t'}, {T'}_{t'})$ from $M$
    %\STATE compute $y_{targ}$ of $Q(s,a; \theta_{targ})$ on ${r'}_t, s_{t'}, t_{t'}$
    \STATE $Q_{targ} \gets {r'}_t + {T'}_{t'} \cdot \gamma \cdot \max_a Q({O'}_{t'}, a ; \theta_{target})$
    \STATE update $\theta$ via gradient descent on $(Q({O'}_t,{a'}_t; \theta) - Q_{targ})^2$
    %\STATE // \emph{Reset environment as needed}
    %\IF {$t_{t'}$ is $True$}
    %    \STATE \texttt{env.reset()}
    %\ENDIF
    %\STATE // \emph{Periodically update theta\_targ}
    \IF {\texttt{step} $\% 1000 == 0$}
        \STATE $\theta_{target} \gets \theta$
    \ENDIF
\ENDWHILE

\end{algorithmic}
\end{algorithm}

\subsection{OEN-based Q-learning}
\label{subsect:OEN-Q}

Motivated by the deep set network of \cite{zaheer2017deep}, we develop an object embedding network to implement our DRL agent, as illustrated in Fig. \ref{fig:OEN}. In this OEN,
the embedding network consists of four layer of $128$ ReLu units, with an ``equivariant" transformation $f_{equiv}(\pmb x, X) = \pmb x - maxpool(X)$ between each layer.
To generate a unified object representation, all embedding representations are pooled by element-wise max pooling across the whole set. The task network consists of three fully connected hidden layers of $128$ ReLu units, and a final output layer of $M$ linear units corresponding to value functions of $M$ possible actions used in playing the given video games. It is worth mentioning that our OEN-based DRL implementation is largely identical to the DQN-based DRL \cite{mnih2015human} apart from two aspects: a) we use the OEN shown in Fig. \ref{fig:OEN}, while the DQN uses a deep CNN as a learning model, and b) our OEN works on object feature vectors, while the DQN works on raw video data. Thus, the same deep Q-learning algorithm can be adapted to train our OEN-based DRL agent.

%Q-learning \cite{watsonQ} is a general reinforcement learning technique that involves keeping track of state-action value estimates $Q(s,a)$ which are predictions of the expected future reward of the agent under the agent's current policy, starting from state $s$ and selecting action $a$. These Q-values are updates as reward samples are collected from the environment, typically bootstrapped based on Q-values of future states. The best policy for the agent is then to select the action which maximises $Q(s,a)$ for the agent's current state.

In a Q-learning based DRL algorithms, the Q-values are estimated using a value function given by a neural network $Q(s,a; \theta)$, where $s$ is a game-state, $a$ is a selected action and $\theta$ is a collective notation of all the parameters in this neural network. The policy for the agent is then given by selecting the action which maximises $Q(s,a; \theta)$ for the agent's current state.
In the DRL learning algorithm proposed by Mnih et al \cite{mnih2015human},
the DQN actually outputs a value vector, $\pmb Q(s; \theta)$, for all the actions simultaneously, where $Q(s, a_m; \theta)$ is the $m$th element of this output vector, reflecting the value of the $m$th action.
Given a sequence $(s_t, a_t, r_t, s_{t+1})$, we can obtain an estimate for $Q(s_t, a_t)$ using the Bellman equation:
$$
    \hat Q(s_t, a_t) = r_t + \gamma \max_a Q(s_{t+1}, a).
$$
This estimate is used as a target $Q_{targ}$ for $Q(s_t, a_t)$. Thus, we define the Q-learning loss function at game-state $t$ as follows:
$$
    L(\mathcal{D}_t; \theta) = (Q(s_t, a_t; \theta) - Q_{targ})^2,
$$
where $\mathcal{D}_t = (s_t, a_t, r_t, s_{t+1})$ is training data retrieved from an experience-replay memory \cite{mnih2015human}. Since the new estimate $Q_{targ}$ depends heavily on the previous values $Q(s_{t+1}, a; \theta)$, a separate network parameterised with the known $\theta_{target}$ is used to obtain $Q_{targ}$ estimates. $\theta_{targ}$ is then updated once the new parameters $\theta$ in the OEN are achieved during the learning. To optimise the loss function for the Q-learning, we employ the Adam optimizer \cite{kingma2014adam}, a gradient-based optimiser. Network parameter update is also done in mini-batches of multiple state-actions and targets simultaneously. Since multiple sequences of different lengths are not readily expressible as fixed-size tensors (which is required by most deep-learning libraries), for each batch we pad each sequence with zero vectors until they have the equal length. A mask of these zero elements is produced and used to nullify their contribution to the output of the OEN. For clarity, we describe the detailed Q-learning algorithm used for training our OEN in Algorithm \ref{alg:objdqn}.

%==================================================

\section{Experiment}\label{sec:experiments}

\begin{figure*}
\vspace*{-5mm}
    \centering
    \begin{subfigure}[m]{4.69cm} % 30
        \includegraphics[width=4.69cm]{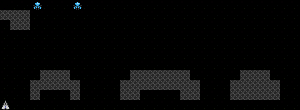}
        \caption{}
        \label{fig:gamesaliens}
    \end{subfigure}
    \begin{subfigure}[m]{4.06cm} % 26
        \includegraphics[width=4.06cm]{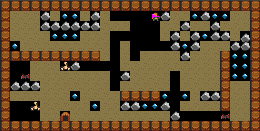}
        \caption{}
        \label{fig:gamesboulderdash}
    \end{subfigure}
    \begin{subfigure}[m]{3.75cm} % 24
        \includegraphics[width=3.75cm]{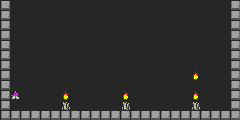}
        \caption{}
        \label{fig:gamesmissilecom}
    \end{subfigure}
    \begin{subfigure}[m]{2.97cm} % 19
        \includegraphics[width=2.97cm]{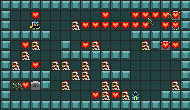}
        \caption{}
        \label{fig:gamessurvivezom}
    \end{subfigure}
    \begin{subfigure}[m]{2.03cm} % 13
        \includegraphics[width=2.03cm]{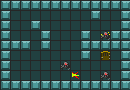}
        \caption{}
        \label{fig:gameszelda}
    \end{subfigure}
    \vspace*{-3mm}
    \caption{Five games from Test Set 1 of the GVG-AI Competition used in our experiments: (a) Aliens. (b) Boulderdash. (c) Missile Command. (d) Survive Zombies. (e) Zelda. Missile Command, Boulderdash and Zelda are based on classic arcade games of the same names, while Aliens is loosely based on the game \emph{Space Invaders}. } \label{fig:games}
\vspace*{-5mm}
\end{figure*}

In this section, we evaluate the performance of our OEN-based DRL agent on five selected games used in the GVG-AI competition \cite{perez2018general}
by comparing with two baseline agents that use different representations of the game-state, and measure average performance during training on a variety of games. To ensure a fair comparison between the different forms of representations, for each agent we change only the agent's neural network and the format of the observation presented to the agent, keeping the rest of the agent design the same.

\subsection{Test environment}

% We use games from GVG-Ai competition
In order to adequately test our approach, we require a corpus of distinct video-games, preferably unified under a single framework.
A common choice for this is the \emph{arcade learning environment} (ALE) \cite{bellemare2012arcade},
however, the ALE does not provide access to ground truth information about object attributes, hence we instead look to the GVG-AI Competition.

The GVG-AI competition \cite{perez2018general} is a regular competition challenging researchers to build AI agents capable of playing a variety of different video-games. These games are specified via \emph{video-game description language} (VGDL) \cite{schaul2013video} where games are defined by a block-based sprite system, and are often based on well-known titles. Importantly, games from previous rounds of the competition are released to the public, proving a large collection of games, all running within the same framework. See Figure~\ref{fig:games} for some examples of games from the GVG-AI Competition.
% either show all five games used in the experiments (if there is space) or remove this figure. For the former, you should make each plot smaller and put all five plots in one row (as same as Fig. 5). Also, the caption should be much more brief such as "Five GVG-AI competition games used in our test. (a) . (b) . (e)." see the caption of Fig. 5 for example.

% Also features a set of adversarial games which can be used blah

%Include Screenshot of games
%\begin{figure}
%    \centering
%    \includegraphics[width=8cm]{BoulderDash.png}
%    \caption{Boulderdash, one of the games from Test Set 1 of the GVG-AI Competition. Like many other games in the GVG-AI Competition, it is based on a classic arcade game of the same name. }
%    \label{fig:boulderdash}
%    %TODO: add images of other games
%\end{figure}

Importantly, information about the current game-state is presented to the agent in the form of a state-observation which includes a list of information about the various sprites (i.e., game objects) within the game. Additionally, while not explicitly provided to the agent, a video output is also produced for human consumption.

%Due to the nature of the competition (agents are given access to a forward-model of the game), most of the research surrounding the GVG-AI competition has focused on search-based AI (also known as planning agents), with only a couple of proposals for learning-based agents to date \cite{samothrakis2015neuroevolution,kunanusont2017general}.

As well as giving direct access to in-game objects, VGDL provides a number of other benefits as a reinforcement learning test-bed, including:
\begin{itemize}
    \item Existing tasks can easily be modified to test how this affects the agent.
    \item New games/tasks can be quickly synthesised for particular purposes.
    \item A large available pool of pre-existing tasks in the form of GVG-AI competition games.
    \item Direct access to the underlying mechanics and ontology, which may be useful as a ground-truth for investigating things such as model-based agents and transfer learning.
\end{itemize}

%Although, potential criticisms of using this framework for testing agents include that successful agents may be overfit to the limitations of the framework, and there is a temptation to select (and/or synthesise) which accentuate the strengths of a particular agent. To tackle these problems, we suggest the creation of a standardised test set of games which reflect a wide variety of different game types. For our purposes, we use the 10 games given in the first GVG-AI competition.

%While the VGDL library used by the GVG-AI Competition currently supports more games than the original VGDL implementation (\texttt{py-vgdl}), we base our environment off of \texttt{py-vgdl} % as the competition version is written in Java, whereas most common deep learning libraries use Python.
%as the current client for the GVG-AI Competition does not currently support image observations.
%We adapted \texttt{py-vgdl} to work with the Open-AI Gym, adding in some extra functionality from the competition version such as support for sprite images.\footnote{Our environment code is available at: \url{https://github.com/EndingCredits/gym_vgdl}}%\footnote{In 2017 a learning track of the competition was introduced, including support for python agents. However this did not include support for image representations.}

%\footnote{While we use the most up-to-date game descriptions possible, there still remain some differences between our environments and the \emph{Java} GVG-AI Competition version, meaning that our results may not be directly comparable with experiments done with the Competition environment.}

For our environment we adapted the original \texttt{py-vgdl} code \footnote{Available here: \url{https://github.com/EndingCredits/gym_vgdl}}, adding in some extra functionality from the competition version such as support for sprite images. Another option would have been to use the code provided for the GVG-AI Competition, although when we started this work there was no python client available.

\begin{figure*}
\vspace*{-5mm}
    \centering
    \begin{subfigure}[m]{6cm}
        \includegraphics[width=6cm]{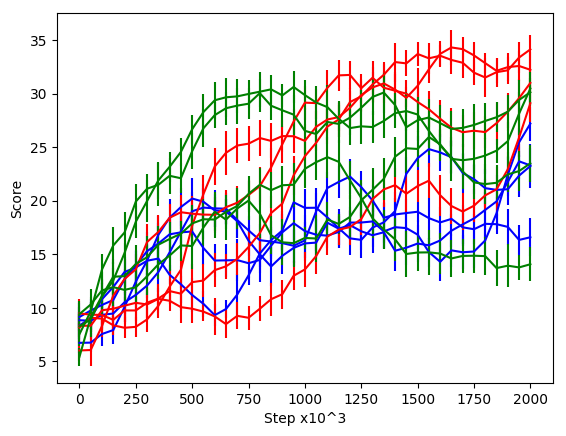}
        \caption{}
        \label{fig:resultsaliens}
    \end{subfigure}
    \begin{subfigure}[m]{6cm}
        \includegraphics[width=6cm]{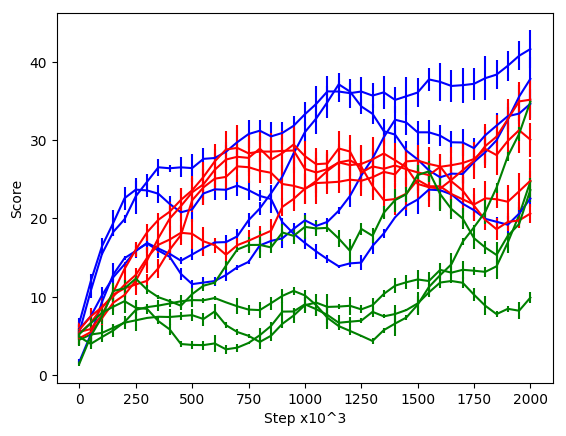}
        \caption{}
        \label{fig:resultsboulderdash}
    \end{subfigure}
    \begin{subfigure}[m]{6cm}
        \includegraphics[width=6cm]{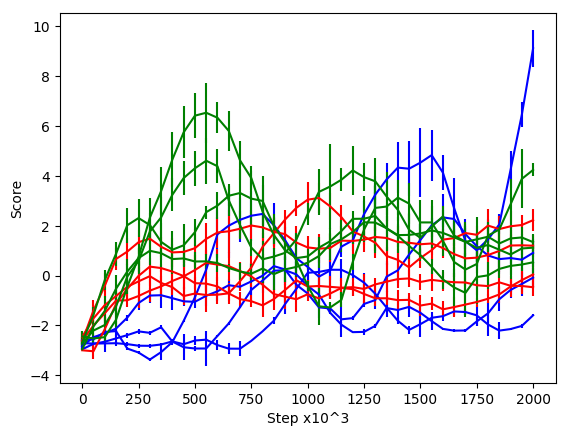}
        \caption{}
        \label{fig:resultsmissilecom}
    \end{subfigure}

    \begin{subfigure}[m]{6cm}
        \includegraphics[width=6cm]{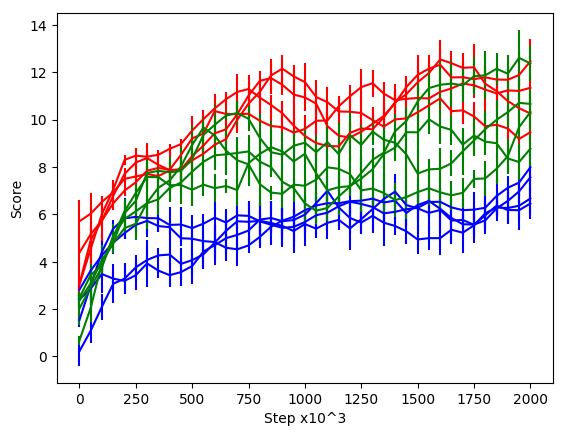}
        \caption{}
        \label{fig:resultssurvivezom}
    \end{subfigure}
    \begin{subfigure}[m]{6cm}
        \includegraphics[width=6cm]{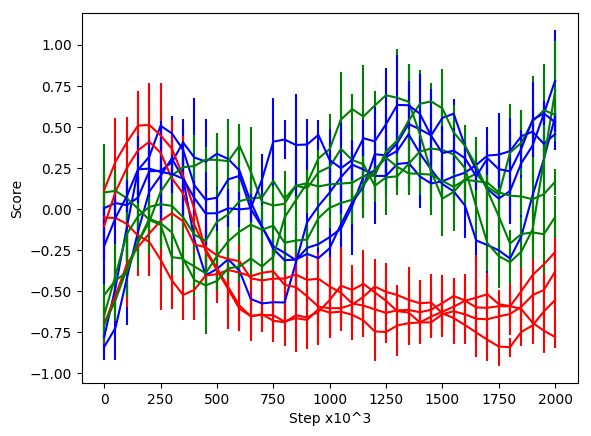}
        \caption{}
        \label{fig:resultszelda}
    \end{subfigure}
    \vspace*{-3mm}
    \caption{Average episode rewards over 50 episodes at various points during training for five games: (a) Aliens. (b) Boulderdash. (c) Missile Command. (d) Survive Zombies. (e) Zelda.
    Results are are smoothed using a forth-order Savitzky-Golay filter with a window size of 21 to improve readability. Lines in Blue are agents with feature representation \& fully connected network, Red are image representation \& CNN, and Green are object representation \& OEN (ours). Best viewed in colour.} \label{fig:results}

    %TODO: update to most recent results
    %TODO: Add baseline score of Random-agent
\vspace*{-5mm}
\end{figure*}

\subsection{Experimental Settings}

To evaluate the performance of each agent we train the agent for 2,000,000 steps, testing the agent for 50 episodes (without training) every 50,000 steps, and record the reward obtained over each of these episodes. While there are a number of different indicators of agent performance, e.g., percentage of games won, we select average episode rewards as this most closely reflects the reinforcement learning objective of the agent, hence is less sensitive to factors such as a mis-specified reward. Additionally,  different users may have different criteria for how they want agents to perform during training, i.e., some may be interested in short-term performance after a certain number of training steps, while others may be interested only in the `asymptotic' final performance, hence we record agent performance throughout training.

We modified our environment code to included support for three different forms of observation types:
\begin{enumerate}
    \item Image representation: A sequence of raw pixel images of the game screen, appropriately sized to be close to the $84 \times 84$ post-processed resolution of the original DQN algorithm.
    \item Object representation: A list of game objects, each given in the form of vector with: a one-hot vector of object class, object co-ordinates, object orientation, and values of given object resources.
    \item Feature representation: A list of the shortest distances from each object class to the player avatar, plus a list of any additional avatar resources. This is the same as the features described in \cite{samothrakis2015neuroevolution}.
\end{enumerate}
In order to simplify our input space, and since certain objects in the game are irrelevant (or invisible) to the agent, for each game we defined a list of the important object classes, and ignore any objects from classes not on that list for both the objects and features representation.
We also remove the frame-skip functionality from our agent as GVG-AI Competition games have a slower update rate, so being able to select actions at each step is important.

For each of these three observation types, we modify our baseline agent as follows, giving us three different agents\footnote{Our agent \& experimental code is available at: \url{https://github.com/EndingCredits/Object-Based-RL}}:
\begin{enumerate}
    \item For the image representation we use the same CNN as used in \cite{pritzel2017neural} with a final linear layer of $M$ outputs. Similar to the original DQN algorithm we also compile states from two consecutive frames (we do not use four frames as sprites are visible every frame in VGDL which is not the case for certain ALE games\footnote{Due to the limited sprite buffers of the \emph{Atari} console, a common optimisation is to draw certain sprites only every other frame.}, and this reduces computational burden).
    \item For the object representation we use the OEN described in Figure~\ref{fig:OEN}.
    \item For the feature representation, we use a simple fully-connected neural network with layers of $64$, $64$, $128$, and $M$ ReLu units, respectively.
\end{enumerate}
Where $M$ is the number of possible actions in the given environment. All the agent hyper-parameters use in our experiments are as follows: 
$\gamma$=0.99, $\epsilon$-start=0.5, $\epsilon$-final=0.1, $\epsilon$-anneal-step=500,000, replay-memory-size=50,000 learning-rate=0.00025, mini-batch-size=32, the agent is trained every four steps, and the parameters are initialised randomly with a Gaussian distribution: $\mathcal{N}(0,0.1)$.
% Note: if a table only has one row/column, it is normally inappropriate. Instead, one always give the information in the main text.

We select five games from Test Set $1$\footnote{Found here: \url{http://www.gvgai.net/training_set.php?rg=1}} of the GVG-AI competition as our test set: Aliens, Boulderdash, Missile Command, Survive Zombies, and Zelda. For each of these we use the first level (i.e., \texttt{level\_0}) as our game environment.
We train our agents on each game four times, using a different seed for agent initialisation each time. Initial environment seeds are reset to the same value for each agent, ensuring that there are no differences in agent performance due to different environment initialisations.

\subsection{Results}

\begin{table}[]
    %\vspace*{5mm}
    \centering
    \begin{tabular}{c | c c c c c }
        & \tiny{Aliens} & \tiny{Boulderdash} & \tiny{Missile Command} & \tiny{Survive Zombies} & \tiny{Zelda} \\
        \hline
        Image          & $35.98$ & $41.90$ & $5.72$ & $\pmb{16.40}$ & $1.16$ \\
        Features       & $29.82$ & $\pmb{43.34}$ & $9.12$ & $10.42$ & $1.58$ \\
        \footnotesize{Objects (ours)} & $\pmb{37.30}$ & $35.90$ & $\pmb{10.44}$ & $14.88$ & $\pmb{1.68}$ \\
    \end{tabular}

    \caption{Best mean score for each agent over $50$ episodes.}
    \label{tab:results}
%\vspace*{-6.1mm}
\end{table}

Full results of our experiments are shown in Figure~\ref{fig:results}. We also report the best mean test score on each game for each agent in Table~\ref{tab:results}, as these give an idea of the theoretical max performance of each agent type accounting for variability in agent parameters (although clearly these results are subject to sample bias, and are likely to be overestimates).

% Some notes on the following could be added to the previous sections
% Evaluating the performance of RL agents can be challenging
% - variance in individual episode performance
% - variance in different agent runs
% - instability
% - When to measure performance
% We don't really check long term performance\cdots

Due to the unpredictable nature of deep reinforcement learning we observed a large variance in agent performance between episodes but also between the average of different tests, making it difficult to compare individual agent results.
Additionally, due to time and computational constraints we were only able to train for two million steps (comparable to eight million frames with frame-skip). This is significantly fewer than the tens and hundreds of millions of frames which many agents from the literature are trained for, meaning the results reported here may only be representative of the early stages of training.
Nevertheless, we do observe certain patterns. In particular, there are notable difference in performance between representations. This is not surprising, as it is well known that choice of representation and network architecture has a big impact on performance across other areas of deep learning.
However, this difference in performance is not consistent across all games; different games seem to favour different representations. Surprisingly, the ``features" baseline outperforms each other agents on certain games, despite often obscuring information about the game state (for example, the agent is given distances to certain objects, but not the direction to them, or the number of them).

It is observed from all the experimental results reported above that our object-based agent is capable of learning in all five games we tested it on. Additionally, across all the games, our agent performs comparably with the other two approaches. To this end, our experiments demonstrate that our OEN-based DRL agent can be an effective alternative to the existing agents for playing general video games.

%TODO: Extra visualisation, and experiments showing extra benefits of the agent.

%==================================================

\section{Discussion}
\label{sec:discussion}

%In this section, we discuss the issues and implications arising from our work.

%% Regarding object and feature extraction%%
%%%%%%%%%%%%%%%%%%%%%%%%%%%%%%%%%%%%%%%%%%%%%%%%%%%%%%%%%%%%
While we believe out method is generally widely applicable, a fundamental assumption made for our approach is that the game-state can be expressed in terms of objects.
In some games object information may be unavailable, thus our approach cannot be applied in those games. Nevertheless, our object-based approach could work on those games, e.g., Starcraft,  where image-based approaches, e.g, DQN \cite{mnih2015human}, are difficult to apply due to unavailability of a complete visual-output snapshot of the full game-state (e.g. due to presence of a controllable camera, or mouse-over options).
In general, object extraction from a game-state can be done either directly via access to the ground-truth of the environment or from video or some other sources. When the ground-truth of the environment is available, object extraction is usually straightforward by utilising the object-oriented nature of most games (although this may also rely on certain domain knowledge to identify which objects are relevant to the agent).
Otherwise, objects could be extracted directly from video data based on the state-of-the-art semantic image segmentation techniques. As a game environment is usually much simpler than natural images, existing semantic image segmentation techniques should be sufficient for this task.
Another limitation of our approach is that it requires the user to find a fixed number of attributes to form feature vectors for all objects applicable to the OEN. In future this requirement for a fixed number of attributes across all object classes could be removed by pre-embedding all objects into a fixed-size space, or using class-specific embedding functions.

%% Regarding exploiting contextual information in an OEN%%
%%%%%%%%%%%%%%%%%%%%%%%%%%%%%%%%%%%%%%%%%%%%%%%%%%%%%%%%%%%%
The use of relational information between objects is important when expressing the game-state. Indeed, Liang et.al. \cite{liang2015state} showed that simple relationships between objects form a good feature set for reinforcement learning in video-games.
In our work, we use only a simple method to exploit contextual information which has limited ability to capture these relations.
%in this case, the last paragraph in Sect. III should be the 1st sentence of this paragraph to explain what contextual information means.
In set networks, there are more sophisticated methods to exploit the contextual information, e.g., those used in \cite{santoro2017simple, watters2017visual}. To achieve more effective unified object representations, our OEN model described in Section \ref{subsect:OEN-Q} can be improved by adopting those techniques developed for set networks.

%% Regarding extension of OEN-based agent beyond DQN%%
%%%%%%%%%%%%%%%%%%%%%%%%%%%%%%%%%%%%%%%%%%%%%%%%%%%%%%%%%%%%

%The Q-learning algorithm used for training our OEN is largely identical to that for the DQN developed by \cite{mnih2015human}. However, a variety of DRL algorithms, e.g., \cite{mnih2016asynchronous, pritzel2017neural} %schulman2015trust,
%can also be applied to train our OEN-based agent in a similar fashion. Moreover, the unified object representation produced by our OEN leads via learning paves a new way to express game-states. Thus, we believe that such a generic game-state representation can be incorporated into not only DRL but also other techniques, such as neuro-evolution and imitation learning, in creating agents for general video-game playing.

% Conclusion%%
%%%%%%%%%%%%%%%%%%%%%%%%%%%%%%%%%%%%%%%%%%%%%%%%%%%%%%%%%%%%
To conclude, we have presented a novel approach to learning directly from semi-structured object information via an OEN for playing general video games. A comparative study based on five GVG-AI competition games suggests that our approach yields performance comparable to two state-of-the-art approaches in general. In our ongoing research, we aim to address the issues and the limitations discussed above for improvement.

%==================================================

%\section*{Acknowledgement}

%The authors would like to thank\cdots

\appendices

\ifCLASSOPTIONcaptionsoff
  \newpage
\fi


\bibliographystyle{IEEEtran}
\bibliography{ReinforcementLearningTheory,VideogameAI,Misc}

%\begin{biography}[{\includegraphics[width=1in,height=1.25in,clip,keepaspectratio]{mshell}}]{Michael Shell}
%\begin{IEEEbiography}[{\includegraphics[width=1in,height=1.25in,clip,keepaspectratio]{picture}}]{John Doe}
%\blindtext
%\end{IEEEbiography}


%\vfill


\end{document}